%% file: root.tex
\documentclass[letterpaper, 10 pt, conference]{ieeeconf}  % Comment this line out if you need a4paper

\IEEEoverridecommandlockouts                              % This command is only needed if 
\usepackage{graphicx}
\usepackage{amsmath,amssymb} % define this before the line numbering.
\usepackage{color}

\usepackage{subfig}

\usepackage{algorithm}
\usepackage{algorithmic}

\usepackage{bm}
\usepackage{todonotes}
\usepackage{color}
\usepackage{multirow}
\usepackage{enumerate}

\newcommand{\eg}{{{\em e.g.}}}

\newcommand{\scz}{\scriptsize}
\title{\LARGE \bf
OmniDRL: Robust Pedestrian Detection using \\ Deep Reinforcement Learning on Omnidirectional Cameras* \\}

\author{G. Dias Pais$^{1}$, Tiago J. Dias$^{1}$, Jacinto C. Nascimento$^{1}$, and Pedro Miraldo$^{2}$% <-this % stops a space
\thanks{*This work was supported by the FCT projects UID/EEA/50009/2019 \& PTDC/EEI-SII/4698/2014, and grant SFRH/BPD/111495/2015. P. Miraldo was partially supported by the Swedish Foundation for Strategic Research (SSF), through the COIN project.}.% <-this % stops a space
\thanks{$^{1}$G. Dias Pais, T. J. Dias, and J. C. Nascimento are with the Institute for Systems and Robotics (LARSyS), Institute Superior T\'{e}cnico, Universidade de Lisboa, PORTUGAL. E-Mail;
{\tt\small goncalo.pais@tecnico.ulisboa.pt}, and {\tt\small \{tdias,jan\}@isr.tecnico.ulisboa.pt}.}%
\thanks{$^{2}$P. Miraldo is with the Division of Decision and Control Systems, KTH Royal Institute of Technology, Stockholm, SWEDEN. E-Mail: {\tt\small miraldo@kth.se}.}%
}

\begin{document}

\maketitle
\thispagestyle{empty}
\pagestyle{empty}

%%%%%%%%%%%%%%%%%%%%%%%%%%%%%%%%%%%%%%%%%%%%%%%%%%%%%%%%%%%%%%%%%%%%%%%%%%%%%%%%
% Abstract
\input{files/abstract.tex} % Introduction
\input{files/intro.tex}    % Background
\input{files/omni.tex}     % Omnidirectional vision
\input{files/deep2.tex}    % Deep Reinforcement Learning
\input{files/experiments.tex} % Experiments 
\input{files/conclusion.tex}  % Conclusions

%%%%%%%%%%%%%%%%%%%%%%%%%%%%%%%%%%%%%%%%%%%%%%%%%%%%%%%%%%%%%%%%%%%%%%%%%%%%%%%%

% \addtolength{\textheight}{-12cm}  % This command serves to balance the column lengths

%%%%%%%%%%%%%%%%%%%%%%%%%%%%%%%%%%%%%%%%%%%%%%%%%%%%%%%%%%%%%%%%%%%%%%%%%%%%%%%%

%%%%%%%%%%%%%%%%%%%%%%%%%%%%%%%%%%%%%%%%%%%%%%%%%%%%%%%%%%%%%%%%%%%%%%%%%%%%%%%%

% \section*{ACKNOWLEDGMENT}
% 
% The preferred spelling of the word �acknowledgment� in America is without an �e� after the �g�. Avoid the stilted expression, �One of us (R. B. G.) thanks . . .�  Instead, try �R. B. G. thanks�. Put sponsor acknowledgments in the unnumbered footnote on the first page.

\clearpage
%%%%%%%%%%%%%%%%%%%%%%%%%%%%%%%%%%%%%%%%%%%%%%%%%%%%%%%%%%%%%%%%%%%%%%%%%%%%%%%%
\bibliographystyle{IEEEtran}
\bibliography{IEEEabrv,egbib}

\end{document}

%% file: files/abstract.tex
\begin{abstract}
Pedestrian detection is one of the most explored topics in computer vision and robotics.
The use of deep learning methods allowed the development of new and highly competitive algorithms.
Deep Reinforcement Learning has proved to be within the state-of-the-art in terms of both detection in perspective cameras and robotics applications.
However, for detection in omnidirectional cameras, the literature is still scarce, mostly because of their high levels of distortion.
This paper presents a novel and efficient technique for robust pedestrian detection in omnidirectional images.
The proposed method uses deep Reinforcement Learning that takes advantage of the distortion in the image.
By considering the 3D bounding boxes and their distorted projections into the image, our method is able to provide the pedestrian's position in the world, in contrast to the image positions provided by most state-of-the-art methods for perspective cameras.
Our method avoids the need of pre-processing steps to remove the distortion, which is computationally expensive.
Beyond the novel solution, our method compares favorably with the state-of-the-art methodologies that do not consider the underlying distortion for the detection task.
\end{abstract}

%% file: files/intro.tex
\section{Introduction}
\label{sec:intro}

Object detection has been one of the most relevant research topics in the last two decades, in both robotics and computer vision communities.
Along with the availability of larger image datasets ({\em e.g.},~\cite{Rus15,Lin14,Eve12,Mot14}) and the increase of hardware capabilities, based on GPUs, we have observed a significant improvement in detection and classification~\cite{Ren15}.
Nowadays, pedestrian analysis is mainly addressed using Deep Learning (DL) techniques (\eg,~\cite{Kri12,Liu17,Rib17}) which attempt to learn high level abstractions of the data.
One of the most important and well-known techniques in DL are the Convolutional Neural Networks (CNNs)~\cite{Kri12}, which halved the error rate for image classification.
Very recently, there have been some alternative methods, namely the use of deep Reinforcement Learning (RL), where a policy is created in order to maximize the rewards of an agent~\cite{Mni15,Cai15}.
On the other hand, omnidirectional cameras have been used in an increasing number of applications ranging from surveillance to robot navigation~\cite{Nev11,Lou14,Ber16,Zhang16}, which require a wide Field of View (FoV) of the environment. This wide FoV is usually obtained by means of distortion~\cite{Swa03}.

An alternative solution for Pedestrian Detection (PD) using omnidirectional cameras would be to first apply undistortion techniques~\cite{Gey00,Bar01,Mei07}, followed by conventional detection tools.
However, these undistortion procedures suffer from the following shortcomings:
1) They are too computationally expensive;
2) They generate images whose pedestrians cannot fit properly in regular bounding boxes\footnote{Although undistorted images keep the perspective projection constraints ({\em i.e.}, straight lines in the world will be projected into straight lines in the image), the objects will be stretched.
This means that the objects should not be approximated by regular bounding boxes, which are used in most of the DL techniques for object detection (we note that there are alternatives that do not consider regular bounding boxes in~\cite{Jaderberg15}).}; 
and
3) They introduce artifacts in the undistorted image that affects the detection accuracy.

\begin{figure}[t]
\centering
    \includegraphics[height=0.22\textheight]{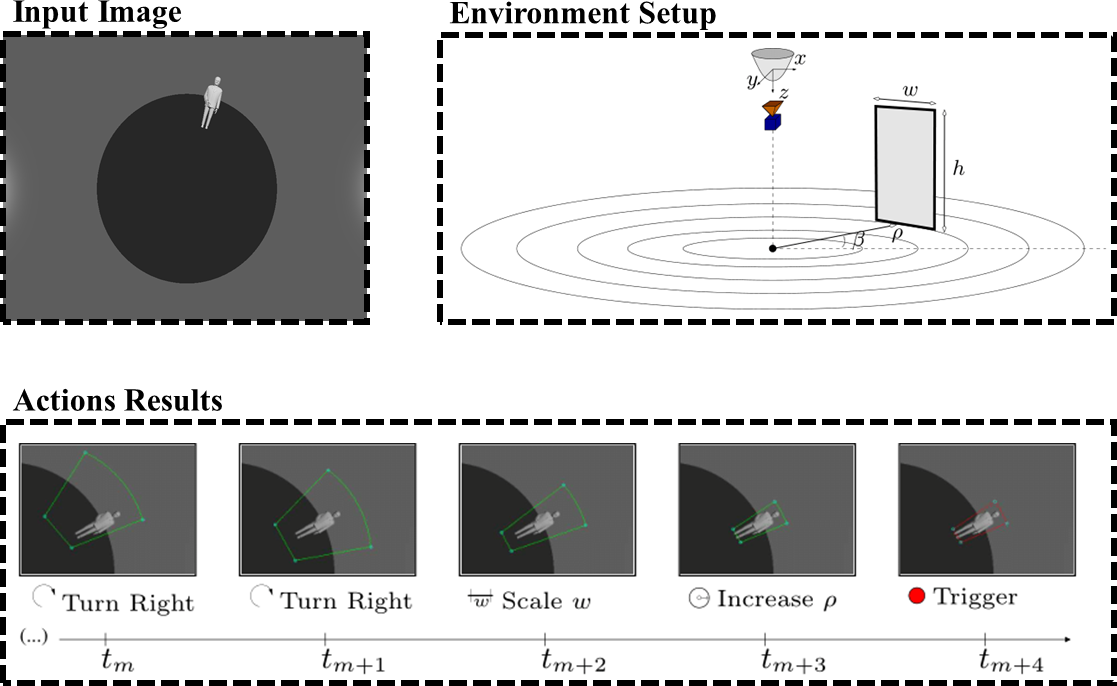}
    \caption{Illustration of the proposal, using a Multi-task network, for pedestrian detection in omnidirectional cameras. The input is an omnidirectional image with an initial state of the bounding box, represented in the world coordinate system. Using this information, a set of possible actions are applied in order to detect the pedestrian in the 3D environment. After the trigger is activated, the line segments of 3D bounding box estimated are projected to the omnidirectional image. Then, the IoU between the ground truth and our estimation is computed in the image coordinates.}
    \label{fig:our_proposal}
\end{figure}

Fig.~\ref{fig:our_proposal} shows the following steps that describe our approach: (i) an omnidirectional image containing the pedestrian (``Input Image''); (ii) our parametrization, using cylindrical coordinates that best represents the 3D environment for omnidirectional cameras (``Environment Setup''); and (iii) the 2D representation of the distorted bounding box states given by the attention mechanism, performed through the agent's actions, for the estimation of the pedestrian 3D position (``Action results'').

\begin{figure}[t]\vspace{1.5mm}
    \includegraphics[height=0.165\textheight]{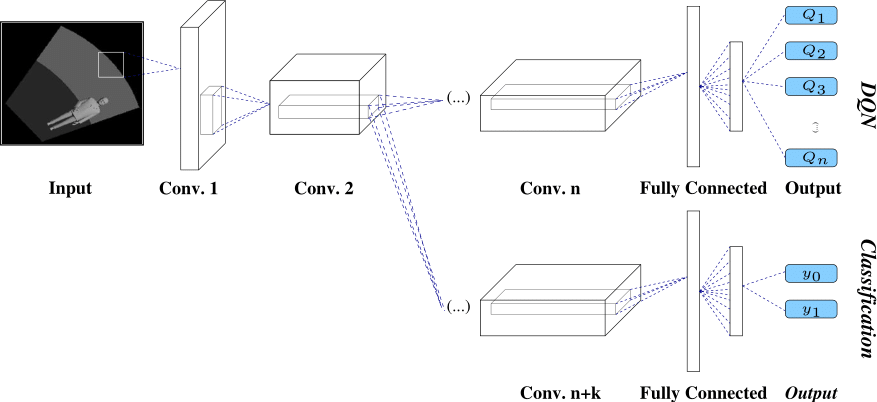}
    \caption[Network Representation.]{Depiction of the scheme of the proposed network, where the first convolutional layers are shared, and then split into branches (DQN and Classification).}
    \label{fig:network_representation}
\end{figure}

\subsection{Related Work}
This section revises related approaches in object/pedestrian detection and omnidirectional vision systems.

\vspace{0.25cm}\noindent{\bf Object Detection and Classification:}
Object detection has a wide range of applications comprising quite diverse research areas, such as artificial intelligence, video surveillance and multimedia systems.
This problem has seen a significant progress in the last decade. 
Looking at the performance of algorithms in, \eg,~PASCAL VOC~\cite{Eve12} datasets, we can easily conclude that the progress slowed from 2010 onward.
However, the use of CNNs~\cite{Kri12} has led to a significant improvement in the accuracy of image classification, for the Imagenet Large Scale Visual Recognition Challenge (ILSVRC)~\cite{Den12}. 
Soon, more methodologies became available, such as: SppNet~\cite{He14}, R-CNN~\cite{Gir14,Hos14}, Mask~\cite{He17}, Fast~\cite{Gir15}, Faster R-CNN~\cite{Ren15_2}, SSD~\cite{Liu16} and YOLOv3~\cite{Red18}.

Although the methods presented above are valuable contributions in the field, we follow a different approach based on deep RL.
One of the reasons behind is that deep RL can significantly reduce the inference time for object detection, when compared to an exhaustive search~\cite{Cai15,Mai17}.
In~\cite{Cai15} it is proposed the use of a Deep Q-Network (DQN)~\cite{Mni15} for object detection, enabling the use of a small amount of training data without compromising the accuracy.
The Q-Network has also been used in other domains, such as medical imaging analysis, \eg,~\cite{Ghe16,Mai17}.
However, the above frameworks have not been explored to model images containing objects/pedestrians with high distortion, such as the ones encountered in omnidirectional systems.

Since the classification task is highly dependent on the bounding box proposal, we propose a {\em  multi-task} network that learns both tasks at the same time (see Fig.~\ref{fig:network_representation}). It will be experimentally shown that the proposed methodology reduces the computational time and increases classification and detection accuracy.
In our framework a hard\footnote{The term hard means that the first convolution layers are shared, and then a branch is created for each sub-network, see Fig.~\ref{fig:network_representation}.} parameter sharing network~\cite{Bax97} is used.
In the {\em multi-task} methodology, each sub-network is trained alternatively.

\vspace{0.25cm}\noindent{\bf Omnidirectional Vision:}
Omnidirectional cameras achieve a wide FoV from two distinct camera configurations: 1) Using special types of lenses (dioptric cameras), such as fish-eye \cite{Kan06}; and 2) Combining quadric mirrors with perspective cameras (catadioptric cameras) \cite{Nay97,Bak99,Mir18}.
Image formation for these systems has been largely studied in the literature.
\cite{Bak99} describes the necessary conditions to ensure that a catadioptric system is a central camera.
In general, these are non-central~\cite{Swa03}.
An alternative way to build an omnidirectional camera would be a setup with multiple perspective cameras.
However, this requires finding correspondences between features and merging images from different cameras. This requires a lot of computational effort, while merging images from different views, details and properties of the environment are lost~\cite{Sch01}.

A small amount of research addressed the object/person detection in omnidirectional images.
One work that is related to our approach is presented in~\cite{Cinaroglu16}. The authors adopt a conventional camera approach that uses sliding windows and Histogram of Gradients (HOG) features.
Since the shape of the sliding window depends on the position of the person w.r.t. the camera, the HOG filters are trained with perspective cameras and then changed to account for the distortion.
Although this method does not require the image unwrapping, avoiding expensive computational resources, it suffers from artifacts caused by changes in the resolution.
By introducing the deep RL in omnidirectional images (OmniDRL), we aim at avoiding the above shortcomings.

There have been some advances in feature extraction for omnidirectional systems (\eg,~\cite{Lou12}). However, this is still a difficult task to be accomplished due to the distortion in the image.
The shape of the features depend on their position on the image (due to the distortion). This is why the traditional feature extraction methods are not suited for these systems.
Some methods exist for soccer robots using catadioptric systems~\cite{Nev11}. However, they require the knowledge of the object shape and color.
There are also tracking methods for omnidirectional cameras based on object's motion, by using background subtraction~\cite{Yam03}, and its ego-motion~\cite{Gan05}. 

Other works address this problem by first applying transformations to the images, followed by the use of conventional techniques, \eg,~\cite{Iraqui10}.
In~\cite{Tang11}, the authors search for objects directly in omnidirectional images, but do not consider the underlying distortion.
In~\cite{Furnari17}, the robustness of affine co-variant features as a function of the distortion is analyzed.

\subsection{Contributions}

We revisit the deep RL in the new context of pedestrian detection in highly distorted scenarios, namely using omnidirectional cameras. The problem formulated herein considers solutions that implicitly take into account distortion's effects on the detection. Our main contributions are as follows:
\begin{enumerate}
    \item Introduction of a novel PD in omnidirectional cameras, inspired in the Deep Q-Network (DQN) methodology and classification tasks. The proposed approach comprises an artificial agent ({\em i.e.}, DQN agent) that can automatically learn policies directly from high dimensional data, to perform  actions in the environment that maximizes the cumulative reward.
    \item Proposal of a novel {\em multi-task} learning strategy that contains two sub-networks: (i) a  DQN, based on deep Reinforcement Learning (see top branch in Fig.~\ref{fig:network_representation}); and (ii) a second classification network which provides a {\em prior} for the pedestrian localization that helps the job of the DQN sub-network (see bottom branch in Fig.~\ref{fig:network_representation}). 
    \item Opposing to the related methods, where the actions are performed in the image domain, our approach is able to perform actions in the world coordinate system. This allows us to know the location of the pedestrian in the 3D environment, {\em i.e.}, the pedestrian's position relative to the camera system can be obtained. This is a useful feature in human-awareness navigation \cite{Rib17,Mat16}.
\end{enumerate}

%% file: files/omni.tex
\section{Line Segment Projection using Omnidirectional Cameras}
\label{sec:back}

We start by defining the image formation model, Sec.~\ref{sec:image_form}. Then, we describe the projection of 3D line segments, required to define the bounding box projection (Sec.~\ref{sec:3d_line}).

\subsection{Image Formation}
\label{sec:image_form}

\begin{figure}[t]
\centering
\subfloat[Image Formation]{\includegraphics[width=0.26\textwidth]{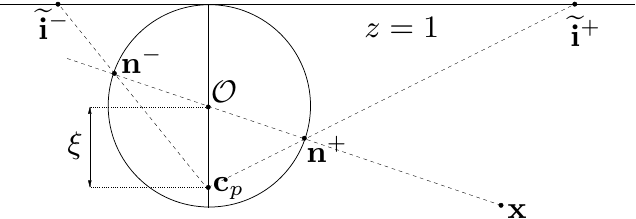}\label{fig:sphereProj_points}}
\subfloat[Projection of 3D Lines]{\includegraphics[width=0.22\textwidth]{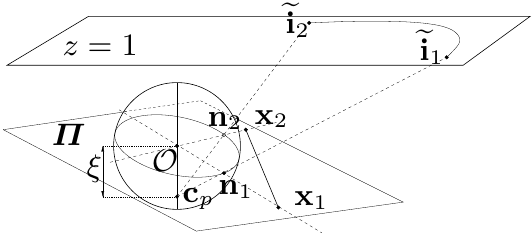}\label{fig:sphereProj_lines}}
\caption[Image Formation for 3D points and 3D lines.]{This figure shows the image formation using unified central catadioptric cameras. In~\protect\subref{fig:sphereProj_points}, it is shown the projection of a point $\mathbf{R}\in\mathbb{R}^3$ onto the normalized image plane $\{\widetilde{\mathbf{i}}^-,\ \widetilde{\mathbf{i}}^+\}$ (in which there is an intermediate projection on the unitary sphere $\{\mathbf{n}^-,\ \mathbf{n}^+\}$).~\protect\subref{fig:sphereProj_lines} shows the projection of 3D straight line segments for images using this model ($\mathbf{x}_1$ and $\mathbf{x}_2$ are the edges of the line's segment).}
\label{fig:sphereProj}
\end{figure}

To deal with general omnidirectional cameras, it is used the spherical model~\cite{Gey00,Bar01,Mei07}\footnote{Notice that, this model can also be used to represent fisheye cameras~\cite{Yin04,Mei07}.}.
Let us assume a unit sphere centered at the origin of the mirror's reference frame. A 3D point $\mathbf{x}= (x_1, x_2, x_3)\in\mathbb{R}^3$ is projected onto the sphere's surface (point $\mathbf{n}\in\mathcal{S}^2\subset\mathbb{R}^3$), resulting in a pair of antipodal points $\left\{ \mathbf{n}^{+}, \mathbf{n}^{-} \right\}$ (see Fig.~\ref{fig:sphereProj}\subref{fig:sphereProj_points} for more details).
The antipodal point closer to $\mathbf{x}$ ({\em i.e.}, $\mathbf{n}^+$ as the convention in the figure) is chosen to be the point projected to the image.

Afterwards, the reference frame is changed and it will be centered at $\mathbf{c}_p = (0,0,\xi)$. In the new reference frame, the point $\mathbf{n}^+$ will be given by the function $\mathcal{H}: \mathbb{R}^3 \rightarrow \mathbb{R}^3$:
\begin{equation}
    \mathbf{n}_{p}^+ = \mathcal{H}\left(\mathbf{x}, \xi\right) = \left(\frac{x_1}{r}, \frac{x_2}{r}, \frac{x_3+\xi}{r} \right),
\end{equation}
where $r = \sqrt{x_1^2 + x_2^2 + x_3^2}$.
Then, $\mathbf{n}^+$ will be projected to the normalized image plane $z = 1$ mapped by the function $\mathcal{F}: \mathbb{R}^3 \rightarrow \mathbb{R}^2$:
\begin{equation}
    \widetilde{\mathbf{i}} =\widetilde{\mathbf{i}}^+ = \mathcal{F}\left(\mathbf{x},\xi\right) = \left(\frac{n_{p,x}^+}{n_{p,z}^+}, \frac{n_{p,y}^+}{n_{p,z}^+} \right). \label{eq:projection}
\end{equation}
Finally, the resulting point has to be mapped on the image plane through the camera projection:
\begin{equation}
\label{eq:generalisedcameramatrix}
    \mathbf{i} = 
    \begin{bmatrix}
        f_1\eta & f_1\eta\alpha \\
        0 & f_2\eta 
    \end{bmatrix}
    \widetilde{\mathbf{i}} + 
    \begin{bmatrix}
        u_0 \\
        v_0
    \end{bmatrix},
\end{equation}
where $f_1$ and $f_2$ are the focal lengths, $u_0$ and $v_0$ are the coordinates of the principal point in the image, and $\alpha$ is the skew parameter~\cite{Har03}.
Finally, $\eta$ is the distance from the center of the first reference frame $\mathcal{O}$ to the image plane.
From~\eqref{eq:generalisedcameramatrix}, one can see that the projection of a point to the image plane is a function of the parameters $\xi$ and $\eta$, which depend on the omnidirectional camera parameters.

\subsection{3D Line Segment Projection}
\label{sec:3d_line}

In central omnidirectional cameras, 3D straight lines are parameterized by a plane $\bm{\Pi}$ (interpretation plane~\cite{Har03}), which comprises by the center of the projection sphere ($\mathcal{O}$ in Fig.~\ref{fig:sphereProj}\subref{fig:sphereProj_lines}) and the respective 3D straight line (notice that, without lost of generality, all lines on this plane will be equally parameterized). Let us denote a line as $\mathbf{l}= (l_1, l_2, l_3)\in\mathbb{R}^3$.

By definition and considering the above parameterization, for two 3D points on the line (say $\mathbf{x}_1$ and $\mathbf{x}_2$), one can represent the line using $\mathbf{l} = \mathbf{x}_1 \times  \mathbf{x}_2 \in\mathbb{R}^3$ (line's moment which is perpendicular to $\bm{\Pi}$), and every point $\mathbf{x}$ belonging to the line must verify $\mathbf{l}^T\mathbf{x} = 0$.

Thus, following~\cite{Bar01}, the image of a 3D straight line on the normalized plane must also belong to a quadric that is defined by the intersection of the plane $\bm{\Pi}$ and the unit sphere.
See Fig.~\ref{fig:sphereProj}\subref{fig:sphereProj_lines} for more details.
After some algebraic manipulations, one can obtain a quadric equation $\mathbf{x}^T \mathbf{C} \mathbf{x} = 0$, where:
\begin{equation}
  \mathbf{C} = 
  \begin{bmatrix}
    l_1^2 \left(1-\xi^2\right) - l_3^2\xi^2 & l_1 l_2 \left(1-\xi^2\right) & l_1 l_3 \\
    l_1 l_2 \left(1-\xi^2\right) & l_2^2 \left(1-\xi^2\right) - l_3^2\eta^2 & l_2 l_3 \\
    l_1 l_2 & l_2 l_3 & l_3^2
  \end{bmatrix},
  \label{eq:Conic}
\end{equation}
represents the curve in the sphere.
To get a representation of the curve in the image plane, we have to consider the constraint  $z=1$, {\em i.e.}, the projection to the normalized plane.
To conclude, one can define the projection curve (image of the line's segment) into the plane by:
\begin{multline}
  \mathcal{G}(\mathbf{x}_1,\mathbf{x}_2) = 
              \big\{ (\widetilde{i}_{x}, \widetilde{i}_{y})\in\mathbb{R}^2 \ : 
                [\widetilde{i}_{x}\ \widetilde{i}_{y}\ 1 ]^T \mathbf{C}
                [\widetilde{i}_{x}\ \widetilde{i}_{y}\ 1 ]  = 0  \\
                \wedge \ \widetilde{i}_{1,x} < \widetilde{i}_{x} < \widetilde{i}_{2,x}\    \wedge \ \widetilde{i}_{1,y} < \widetilde{i}_{y} < \widetilde{i}_{2,y} \ \big\},
  \label{eq:quadricSolve}
\end{multline}
where $(\widetilde{i}_{1,{x}},\widetilde{i}_{1,{y}}\,)=\mathcal{F}({\bf x}_1,\xi)$ and $(\widetilde{i}_{2,{x}},\widetilde{i}_{2,{y}})=\mathcal{F}({\bf x}_2,\xi)$ (see (\ref{eq:projection}) and also Fig.~\ref{fig:sphereProj}\subref{fig:sphereProj_lines}).
Finally, to the each point that belongs to the curve, a final mapping must be applied to project them into the image plane, which is obtained by applying~\eqref{eq:generalisedcameramatrix}.

%% file: files/deep2.tex
\section{Robust Detection}
\label{chp:deep}

In this section, we present the principles behind the training and inference of the deep RL agent and, the classification network used.
We define the agent's training and how it infers its actions in Sec.~\ref{sec:drl}. Further, in Sec.~\ref{sec:class}, the same process is done for the classification task using CNNs.

\subsection{Deep Reinforcement Learning}
\label{sec:drl}

This section describes how object detection is performed using deep RL. Recently,~\cite{Cai15} have proposed the use of a Q-network as a way to improve the efficiency on the detection of an object. The main challenge is to the adapt our DQN for object detection through movements applied to a bounding box, in order to deal with the high distortions created by the omnidirectional imaging device. In order to have a 3D position of the pedestrian, it was considered a 3D bounding box representation in a 3D environment (Fig.~\ref{fig:our_proposal} - ``Environment Setup'').

The DQN is a deep Q-Network parameterized by ${\bm \theta_t}$ that, for a given state $s_t$ approximates the Action-Value Function $Q(s_t,a_t; {\bm \theta}_t)$ for a set of nine actions, that are function of 
the angle ($\beta$), distance ($\rho$), height ($h$) and width ($w$):   
\begin{equation}
    a_t \in \mathcal{A} = \{\rho^{+}, \rho^{-}, \beta^{+}, \beta^{-}, w^{+}, w^{-}, h^{+}, h^{-}, \sigma \},
\end{equation}

\newcommand{\jojo}{\rule[-.2cm]{0cm}{.4cm}}
\begin{table}[t]\vspace{-2mm}
	\begin{center}
	{
	\setlength{\tabcolsep}{3pt}
		\subfloat[Update Position]{
			\begin{tabular}{|c|c|}
				\hline
				{\bf \scz Action} & {\bf \scz Position}                                \\ \hline \hline
				\jojo \scz $a_t = \rho^{+}$,    $a_t = \rho^{-}$    & \scz \jojo $[\rho +a_t, \beta, z]^T$         \\ \hline
				\jojo \scz $a_t = \beta^{+}$,  $a_t = \beta^{-}$  & \scz \jojo $[\rho,  \beta+a_t, z]^T$  \\ \hline
			\end{tabular}
			\label{tab:update_position}}
			\subfloat[Update Dimension]{
				\begin{tabular}{|c|c|}
					\hline
					{\bf \scz Action} & {\bf \scz Dimension}          \\ \hline \hline
					\jojo \scz $a_t = w^{+}$,  $a_t = w^{-}$  & \jojo \scz $[w +a_t, h]^T$  \\ \hline
					\jojo \scz $a_t = h^{+}$,  $a_t = h^{-}$  & \jojo \scz $[w, h+a_t]^T$   \\ \hline
				\end{tabular}
				\label{tab:update_dimension}}
				}
			\end{center}
				\caption[Update rules for position and dimension.]{Actions considered in the proposed deep RL. The $+,-$ stand for positive or negative updates for depth ($\rho$), rotation ($\beta$) z-scale ($h$) and
		$(x,y)$-scale ($w$). In~\protect\subref{tab:update_position} and~\protect\subref{tab:update_dimension} the dimension and the position remains unchanged, respectively.}
			\label{tab:bb_action}
		\end{table}

The  effects of each action $a_t$ on the bounding box position and dimension is shown in Tab.~\ref{tab:bb_action} (see ``Environment Setup'' in Fig.~\ref{fig:our_proposal}).  
The Q-values given by $Q(s_t,a_t; {\bm \theta}_t)$ represent the maximum discounted future reward of the Bellman equation~\cite{Sut98}. Based on the 3D line's projection defined in the previous section, each state (bounding box projection \eqref{eq:quadricSolve}) can be formally defined by:
\begin{multline}
    \label{eq:state_def}
    \mathbf{s}_t = \mathcal{I}\left(\mathcal{G}(\mathbf{x}_i,\mathbf{x}_j)\right) \in \mathbf{S},\\ \text{with} \ (i,j) \in \{(1,2),\ (1,3),\ (2,4),\ (3,4) \},
\end{multline}
where $\mathbf{x}_i$ for $i=\{1,2,3,4\}$ are the 3D bounding box corners, and $\mathcal{I}(.)$ represents the resulting image cropped within the bounding box limits.

The training process of the DQN follows a deep Q-learning algorithm proposed by Mnih~\cite{Mni15}, where the authors used two main key features. First, the process makes use of a memory $\mathbf{D}_t=\{e_1,...,e_t\}$ that is built with the agent's experiences $\mathbf{e}_t = (s_t, a_t, r_t, s_{t+1})$.
These samples are drawn uniformly from the memory and will be used as a batch to train the prediction network.
Second, a target network containing the parameters $\bm{\theta}_{t}^{-}$ computes the target values that allows the DQN updates.
These values $\bm{\theta}_t^{-}$ are held unchanged and updated periodically.

One shortcoming of the DQN is that the action selection and the corresponding evaluation
tend to use the same Q-values, leading to overoptimistic estimates. To tackle 
the above limitation we use the Double DQN (DDQN -~\cite{Has16}). Therefore, the loss function that models $Q(s_t,a_t;\bm{\theta}_t)$ minimizes the following mean square error on the modified version of the {\it Bellman} equation:
\begin{multline}
L_{drl}(\bm{\theta}_t) =  \\ \mathbb{E}_{(s_t,a_t,r_t,s_{t+1})~\sim~U(\mathbf{D}_t)}\left[(q_j^{DDQN} - Q(s_t,a_t;\bm{\theta}_t))^2 \right],
\label{eq:Loss}
\end{multline}
where:
\begin{equation} 
q_j^{DDQN} = 
r_t + \gamma  Q(s_{t+1}, \arg\max_{a_{t+1}}  Q(s_{t+1},a_{t+1};\bm{\theta}_t); \bm{\theta}^-_t),
\end{equation}
where $U(\mathbf{D}_t)$ represents the batch retrieved from the memory $\mathbf{D}_t$ with uniform distribution $U$.

In the training process, the agent must first explore the environment in order to better understand which actions will maximise its future reward and then exploit those states that leads it to that objective.
We choose a strategy that will balance the two phases {\em i.e.}, 
{\em exploration} and {\em exploitation}. For our training, we use the Softmax action selection method using a Gibbs or Boltzmann distribution~\cite{Sut90}.

For inference, the trained DQN model is parameterized by $\bm{\theta}^{\star}$ learned in~\eqref{eq:Loss} that outputs the action-value function for the current state $s_t$. Formally, the action to follow from the current state is defined by:
\begin{equation}
a^{\star}_t = \arg\max_{a_t} Q(s_t,a_t; \bm{\theta}^{\star}).
\end{equation}

\subsection{Classification Branch}
\label{sec:class}

In this section, we describe how the classification network is trained (see classification branch in Fig.~\ref{fig:network_representation}).
The main goal in a classification task is to assign a class to an object, within a proposed region.
These regions can be represented as: rectangular bounding box~\cite{Ren15_2}, segmented region~\cite{He17} or, as we propose, distorted bounding box.

The purpose is to classify the proposed region as having the pedestrian or not, thus we consider only two classes $y_c = \{0, 1\}$ (No Pedestrian and Pedestrian, respectively).
Assuming one pedestrian per image, we consider that for each image $\mathcal{I}(.)$ a label is assigned.
Then, our primary goal is to create a network, described by the parameters ${\bm \theta}_t$, capable of generating a prediction of the desired class for a hold out test image.
Therefore, it was used a CNN to model this behavior by minimizing the classification error of a batch of labeled images, according to cross-entropy's mean:
\begin{multline}
\label{eq:loss_cls}
L_{cls}({\bm \theta}_t) = \\
\mathbb{E}_{~\sim~U(\mathbf{D}_t)}\left[-y_c \log(p(\widehat{y}_1;{\bm \theta}_t)) - (1 - y_c)\log(p(\widehat{y}_0;{\bm \theta}_t))\right],
\end{multline}
where $y_c$ is the labeled class; $p_i(\widehat{y}_i;{\bm \theta}_t)$ is the computed probability for each class; ${\bm \theta}_t$ are the network's weights and $\widehat{y}_i, \ i \in \{0,1\}$ are its estimates (for each class at iteration $t$); and $U(\mathbf{D}_t)$ is the uniformly retrieved batch of labeled images from the memory used in the deep RL system.
This probability is computed using the Softmax function.% of the network's output.

For inference, it is chosen $i$ that has the highest $p_i(\widehat{y}_i;{\bm \theta}_t)$:
\begin{equation}
\label{eq:clss_infer}
y_c^{\star} = \arg\max_{i} p(\widehat{y}_i;{\bm \theta}^{\star}),
\end{equation} 
where ${\bm \theta}^{\star}$ are the optimal parameters of the model learned in~\eqref{eq:loss_cls}. The obtained classification will provide the initial state position for the deep RL branch (see Sec.~\ref{sec:drl}), in testing.

%% file: files/experiments.tex
\section{Experimental Results}
\label{sec:exp}

This section is organized as follows. First, in Sec.~\ref{sec:multi_task_comp} we compare 
the performance between the DQN (see top branch in Fig.~\ref{fig:network_representation}) and the advantages by coupling a classification 
network (bottom branch in Fig.~\ref{fig:network_representation}) that constitutes our multi-task learning methodology. The second part of the experiments (Sec.~\ref{sec:SotA_comp}) concerns a comparison of the proposed multi-task learning method with the State-of-the-Art: the object detection method proposed by Caicedo~\cite{Cai15}; and the "Faster R-CNN" using a ResNet-101 architecture~\cite{He16}, pre-trained with the PASCAL VOC 2007 and 2012 datasets.

As far as we know, there are no available datasets in omnidirectional settings for PD, as well as the corresponding labels. Thus, we acquired a new one using three different environments in our laboratory with several subjects. With a total of 921 images, $70\%$ of them are used for training and $30\%$ are used for testing. 
These images were obtained with~\cite{webflea3}'s Flea3 camera attached to a hyperbolic mirror, modeled and calibrated by the central catadioptric camera system~\cite{Mei07}.
Our method was developed using {\tt TensorFlow} and {\tt OpenCV}. The dataset are available on the authors' webpage.

\subsection{Comparison between Deep RL and Multi task learning}
\label{sec:multi_task_comp}

The architecture of the DQN network comprises five convolutional layers and two fully-connected layers, where the input is a $224 \times 224$ image size, and the output contains the Q-values for the nine possible actions (see top network in Fig.~\ref{fig:network_representation}).
The multi-task network contains a second classification network, having  
three shared convolutional layers, two more convolutional layers and two fully-connected layers for each branch. Now, in addition to the nine Q-values, the Softmax probability is retrieved for the two classes (presence or absence of the pedestrian), in the classification network.
For the training parameter we considered $\tau = 0.6$, $\alpha = 10$, $C = 15000$ steps and a fixed learning rate of $0.0001$.

For the purpose of test initialization, we choose a fixed $\rho$ closed to the camera center and a fixed dimension $\mathbf{dim}_0$ with a relatively large size.
Afterwards, six values for the $\beta$ parameter are randomly selected in order to cover the whole image domain, aiming to detect some pedestrian location.
For the other values, no detection has been triggered (a maximum number of 100 steps has been used).

For evaluation, a hold-out test sequence is used to compute the average steps required to detect the pedestrian, as well as the average of the Intersection over Union (IoU) over the final detections. Notice that the steps correspond to actions taken and are counted from the first initial position until the detection of the pedestrian has been triggered.  

\begin{table}[t]\vspace{2mm}
\centering
\begin{tabular}{c|c|c|c|c|}
\cline{2-5}
& \scz RMSE $\rho$
& \scz RMSE $\beta$ 
& \scz Error Std $\rho$ 
& \scz Error Std $\beta$ \\ \hline %\hline
\multicolumn{1}{|l|}{\scz DQN} 
& \scz $0.6748$
& \scz $0.0875$
& \scz $\bm{0.4097}$
& \scz $0.0837$
\\ \hline 
\multicolumn{1}{|l|}{\bf{\scz Multi-Task}}  
& \scz $\bm{0.5541}$
& \scz $\bm{0.0316}$
& \scz $0.5506$
& \scz $\bm{0.0318}$
\\ \hline
\end{tabular}
\caption[Errors]{This table shows the position errors of $\rho$ (m) and $\beta$ (rad) for ``DQN'' and ``Multi-task'' approaches. The difference of the position obtained when triggered and the ground truth is computed, and then the Root Mean Square Error (RMSE) and the error Standard Deviation (Std).}
\label{tab:position_errors}
\end{table}

Tab.~\ref{tab:position_errors} shows the obtained errors for the duplet $(\rho,\beta)$ for the two approaches, DQN and Multi-task (DQM \& classification branch).
The proposed solutions exhibit small errors in the detection task in spite of the environment's complexity.
However, by observing the position error $\rho$ and its standard deviation, the errors obtained are larger than it would be expected.
This is justified by the IoU not fully demonstrate the changes on the distortion given by $\rho$, {\em i.e.}, $\mathbf{s}_{gt}$ and $\mathbf{s}_{t+1}$ can be identical ($IoU(\mathbf{s}_{gt},\mathbf{s}_{t+1}) > 0.65$), even though the current state is further away from the ground truth, due to the distortion in central catadioptric systems being radial.
Nonetheless, as we can also examine in Tab.~\ref{tab:position_errors}, the position error of $\beta$ is small, thus small changes in $\beta$ will lower the IoU, for the same reason as before.
The distortion in this kind of system will have a higher effect when the bounding box moves along the radius.

\begin{table}[t]
\centering
\centering
\begin{tabular}{c|c|c|c|c|}
\cline{2-5}
& \scz Average Steps
& \scz Average IoU
& \scz Correct (\%)
& \scz Training Steps \\
\hline %\hline
\multicolumn{1}{|l|}{\scz DQN}  & 
\scz $27.85$ & 
\scz $0.623$  & 
\scz $71.3$ &  
\scz $1.306 \times 10^6$ \\ \hline
\multicolumn{1}{|l|}{\bf{\scz Multi-task}} &
\scz $\bm{21.39} $ & 
\scz $\bm{0.718}$  &
\scz $\bm{86.5}$ & 
\scz $\bm{ 0.936 \times 10^6}$\\ \hline
\end{tabular}
\caption[Methods comparison for the proof of concept.]{This table shows a comparison between the ``DQN'' and ``Multi-task'' methods in terms of the average steps needed to correctly trigger the detection and the average IoU, which evaluates the detection regarding the distorted ground truth. The correct percentage of detection and the raining step ({\em i.e.}, the actions) are also shown.}
\label{tab:our_methods_comparison}
\end{table}

Table~\ref{tab:our_methods_comparison} shows the superiority of the Multi-task approach. This is achieved in both accuracy (better IoU scores) and efficiency (less actions needed for the detection). 
From the above, we can state that the classification branch, helps on the training and allow us to get an initial estimation for testing.
In the next section, we perform the comparison with other related methodologies, and the proposed multi-task network will be termed as \textbf{``Ours''}.

\subsection{Comparison with the State-of-the-Art Methods}
\label{sec:SotA_comp}

In this section, it is performed a comparison between the proposed multi-task framework (``Ours'') within the environment discussed in Sec.~\ref{chp:deep}, against two state-of-the-art approaches using rectangular bounding boxes: 1) Caicedo's method; and 2) a pre-trained Faster R-CNN. For the state-of-the art methods, 
two distinct settings are used: i) distorted images acquired by our omnidirectional imaging device (termed as ``SotA'' and ``Faster'', respectively); and ii) unwrapped images obtained from the original images (in the same way, termed as ``SotAU'' and ``FasterU''). Note that the environment where the actions are performed for these methods are in the image domain while ours is in 3D.

The network ``Ours'', ``SotA'' and ``SotAU'' have five convolutional layers and two fully-connected layers, where the input is a $224 \times 224$ image size, and the output contains the Q-values for the nine possible actions. 
Notice that we can maintain the same dimension of actions as in~\cite{Cai15}, thus we can keep the same architecture to train both methods. 
``Ours'' has two more convolutional layers and two fully-connected layers for each branch
(see Fig.~\ref{fig:network_representation}).
In addition to the nine Q-values output, in the ``Ours'' method, the Softmax probability is also retrieved for the two classes in the classification network.
The training parameters are the same for all the above networks. These three networks are trained from scratch. It is important to mention that we used the Faster R-CNN as a baseline approach, in order to verify how a pre-trained network for perspective images behaves when distorted and unwrapped images are tested. Thus, average and training steps will not be evaluated for this approach. 

%--------------------------------------------------

\begin{table}[t]\vspace{2mm}
\centering
{\setlength{\tabcolsep}{5pt}
\begin{tabular}{c|c|c|c|c|}
		\cline{2-5}
		& \scz Average Steps &
		\scz Average IoU  &
		\scz Correct (\%) &
		\scz Training Steps  \\ \hline %\hline
		\multicolumn{1}{|l|}{\scz SotA\cite{Cai15}}  &
		\scz $62.65$  &
		\scz $0.385$ &
		\scz $83.9$ &
		\scz $0.975 \times 10^6$ \\ \hline
		\multicolumn{1}{|l|}{\scz SotAU\cite{Cai15}}  &  
		\scz $28.464$ & 
		\scz -- &
		\scz $77.2$ &
		$1.385 \times 10^6$\\ \hline
		\multicolumn{1}{|l|}{\scz Faster\cite{He16}}  &
		\scz --         &
		$0.608$ & 
		\scz $28.6$ &  -- \\ \hline
		\multicolumn{1}{|l|}{\scz FasterU\cite{He16}}  &
		\scz --         & 
		\scz -- & 
		\scz $68.6$ &
		\scz-- \\ \hline
		\multicolumn{1}{|l|}{\scz {\bf Ours }} &
		\scz $\bm{21.39} $ &
		\scz $\bm{0.718}$ &
		\scz $\bm{86.5}$ & 
		\scz $\bm{ 0.936 \times 10^6}$\\ \hline
	\end{tabular}
	}
	\caption[Methods comparison for the proof of concept.]{This table shows the average steps in the test sequence to set the trigger correctly for the different methods, the average IoU, that evaluate the distorted image on the distorted ground truth, the correct percentage of detections on the test dataset, and the final training steps. The average IoU is only computed for those methods, because the  projection of the bounding box to the omnidirectional image after its undistortion in ``SotAU'' and ``FasterU'' cannot be obtained.}
    \label{tab:all_methods_comparison}
\end{table}

\begin{figure}[t]\vspace{1mm}
\centering
\includegraphics[height=0.22\textheight]{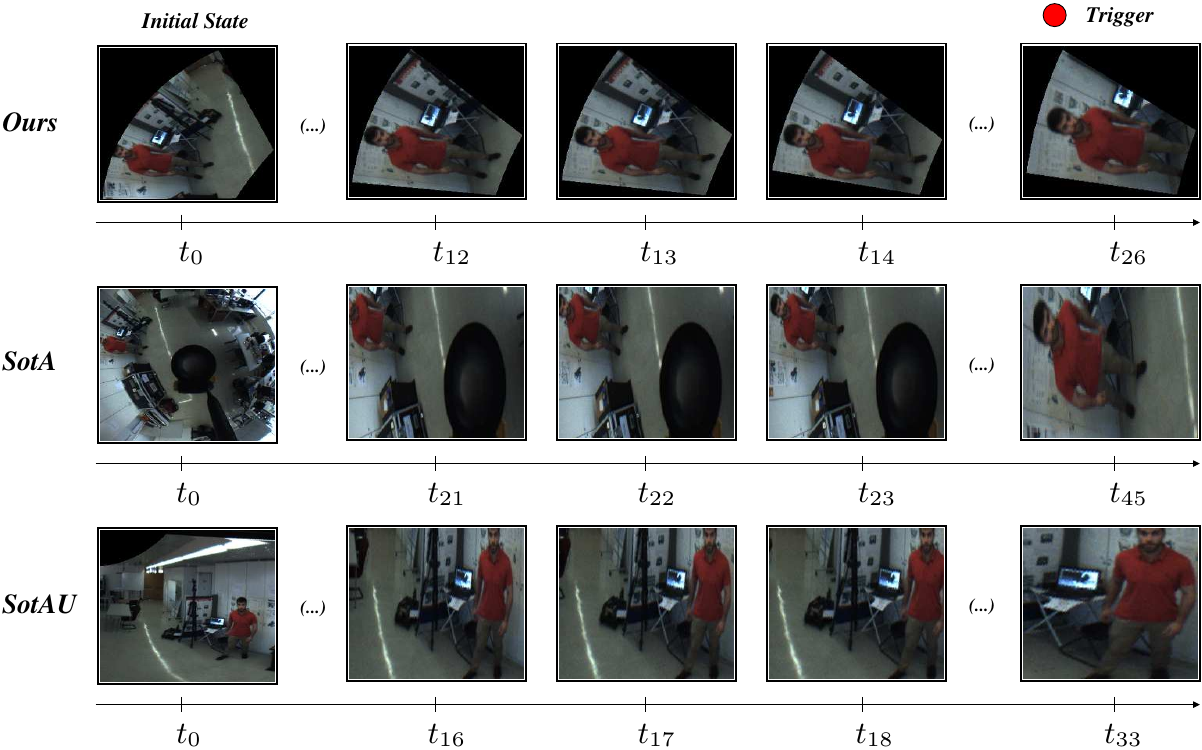}
\caption{Depiction of the temporal evolution on a test image for the three evaluation methods.}
\label{fig:iou_steps}
\end{figure}

For evaluation purposes, we used the test dataset to compute the average steps required to detect the pedestrian, as well as the average IoU for the final detections.
Notice that the steps correspond to actions taken and are counted from the first initial position until the detection of the pedestrian has been triggered.
For the IoU computation, we only take in consideration the three methods that deal with the distorted image ({\em i.e.}, ``SotA'', ``Faster'' and ``Ours'').
From Tab.~\ref{tab:all_methods_comparison}, it can be seen that our method takes much less steps to achieve the correct detection than the ``SotA'', even though the environment is much more complex, as the coordinate system is set in the world coordinates and not in the image domain. An example illustrating the steps required for each algorithm can be seen in the Fig.~\ref{fig:iou_steps}.
Regarding the perspective method on the undistorted image (``SotAU''), we obtained a small number of average steps due to the simplicity of its environment, since the image is much smaller and the initialization of the initial bounding box already covers a big part of the environment.
Also note that our method reaches an IoU improvement of over $33.3\%$ when comparing to the ``SotA'' method and over $11\%$ regarding the ``Faster'' method. However, as it can be seen, the correct detection percentage (shown in Tab.~\ref{tab:all_methods_comparison}) for the ``Faster'' is less $57.9\%$ when compared to our algorithm. One can assume that the high percentage on the average IoU, for the ``Faster'' method, is related with images where the pedestrian is located farther from the imaging device, where the distortion's effects are mostly neglected. 

In the experiments, we consider a correct detection when the trigger is signaled above a threshold\footnote{The same value of $\tau$ is used in training.} or, in the case of the ``Faster'' and ``FasterU'', when a pedestrian is detected in the image. From the Tab.~\ref{tab:all_methods_comparison} we can see that the proposed method ``Ours'' triggers the detection using a smaller number of actions and deals better with the distortion when compared to both ``SotA'' and ``SotAU''.
Our approach outperforms $2.5\%$ points over the previous solutions.
In addition, one can state that the ``SotAU'' method outperforms the ``FasterU'' on the correct detection percentage, since the ``SotAU'' method was trained on images that contain artifacts caused by the unwrapping process while the ``FasterU'' was already pre-trained with images acquired from perspective cameras.

%% file: files/conclusion.tex
\section{Discussion}
\label{sec:concl}

In this work, we have presented a novel methodology for PD in omnidirectional systems.
The approach uses the underlying geometry of general central omnidirectional cameras along with a multi-task learning methodology.
From the extensive conducted evaluation, the methodology is able to accurately detect pedestrians without resorting to undistortion procedures which are computationally expensive. 
Moreover, our proposal can provide the 3D world position of the pedestrian instead of 2D image coordinates.
Also, it needs a smaller number of agent actions to reach the correct detection, and exhibits a higher accuracy on the final bounding box representation.
This is due to the fact that our framework inherently holds high levels of distortions where the bounding box is represented in cylindrical coordinates which are tailored for these omnidirectional environments.